%% file: exp_maxent.tex
\documentclass[fleqn,10pt]{wlscirep}
\usepackage{amsmath}

\title{Maximum entropy models for generation of expressive music}

\author[1]{Simon Moulieras}
\author[1,2]{Francois Pachet}
\affil[1]{SONY CSL, Paris, France}
\affil[2]{Sorbonne Universit\'es, UPMC Univ Paris 06, UMR 7606, LIP6, F-75005, Paris, France}

\affil[*]{simon.moulieras@gmail.com}

\newcommand{\Hcal}{\mathcal{H}}

\newcommand{\Lcal}{\mathcal{L}}
\newcommand{\Zcal}{\mathcal{Z}}
\newcommand{\Dcal}{\mathcal{D}}

\begin{abstract}

In the context of contemporary monophonic music, expression can be seen as the difference between a musical performance and its symbolic representation, \textit{i.e.} a musical score. In this paper, we show how Maximum Entropy (MaxEnt) models can be used to generate musical expression in order to mimic a human performance. As a training corpus, we had a professional pianist play about 150 melodies of jazz, pop, and latin jazz. The results show a good predictive power, validating the choice of our model. Additionally, we set up a listening test whose results reveal that on average, people significantly prefer the melodies generated by the MaxEnt model than the ones without any expression, or with fully random expression. Furthermore, in some cases, MaxEnt melodies are almost as popular as the human performed ones.

\end{abstract}

\begin{document}

\flushbottom
\maketitle

\thispagestyle{empty}

\section*{Introduction}

Human performances of written music differ in many aspects from a straight acoustical rendering a computer would make. Besides the fact that we would not be even able to play music like a machine, a human performer would give its own interpretation to a musical score, by changing the timing, the loudness, the duration, and the timbre of each note, or even by removing or adding notes (in jazz for example). In fact, a human listener can easily recognize a human performance from a straight computer rendering, without necessarily being a musician. Nevertheless, there is no straightforward mathematical formulation of what is or is not musically expressive. 

Musical expression is a fundamental component of music. It can translate the intention of the composer via the indications written on the score, in order to help the performer. These indications can be whether associated to a specific note, or a group of notes (dynamical indications, articulations, ...) or a word corresponding to a specific style, or a specific phrasing (\textit{e.g.} fast swing, ballad, groovy, ...). In the latter case, one can legitimately expect to find non-trivial correlations between some well choosen observables, reflecting the specific musical style, and it is the aim of this paper.

For this purpose we will use a MaxEnt model which belongs to the wide class of probabilistic graphical models, built to capture and mimic both unary and pairwise correlations between variables. MaxEnt models have already been used in music \cite{jason} in order to study melodic patterns. Unlike in Markov chains in which the sampling complexity grows exponentially with the size of the patterns, MaxEnt models consistent with pairwise correlations keep this complexity quadratic, independently from the distance between the correlated variables. Importantly, we will use a translation-invariant model for simplicity, \textit{i.e} we will consider that the probability distribution of a variable associated to a particular note depends only on the variables associated to the neighbouring notes, independently from its intrinsic position in the melody. This assumption is equivalent to saying that musical expression consists in local texture, rather than long-range correlations. 

Like in other aspects of music such as harmony or rhythm, musical expression has a complex underlying structure, that have been tried to be captured by statistical models, or learning algorithm (see \cite{grachten, grachten2012linear} on romantic music). Working with expression differs from symbolic music computing, in the fact that it involves continuous variables like loudness, or local tempo modulation that cannot be mathematically treated as categories (like the pitch, for example). Unlike romantic music in which a performer usually continuously modulate the tempo, contemporary western music like jazz, or pop music is often played with a rhythmic section that tend to maintain a constant tempo along the melody. The rhythmic expression is thus contained in the difference between: \textit{i}: the onset of the performed note and its onset on the score, and \textit{ii}: the duration of the performed note and its duration on the score. In the following, we will refer to these observables as microtiming deviations.

\section{The Model}
\label{sec:model}

\subsection{Construction}
\label{sec:construction}
The Principle of Maximum Entropy (PME) \cite{jaynes1957information} states that given a set of observations, the probability distribution that best model the data is the one that maximizes the entropy. The observations are expressed in terms of averages, or expected values, of one or more quantities. This principle has applications in many domains, such as information theory, biology, statistics, and many more, but comes from statistical physics in which it aimed at connecting macroscopic properties of physical systems to a microscopic description at the atomic or molecular level. 

Given a set of observables $F_j= \langle f_j(x) \rangle|_{P(x)}$, $1\leq j \leq N_c$ on a quantity $x$, in which $\langle \rangle|_{P(x)}$ represents the expectation value over the probability distribution $P(x)$, the PME allows us to determine the distribution $P(x)$ of maximum entropy, fulfilling the constraints. 

The variables we need our model to deal with are: 
\begin{itemize}
\item metrical position in the bar (discrete)
\item onset deviation (continuous)
\item duration deviation (continuous)
\item loudness (continuous).
\end{itemize}

Hence, the PME should be applied for each of these variables, and to carefully treat separately continuous variables that will be denoted by $y$, from discrete ones denoted by $x$. $z$ will refer to an arbitrary variable, belonging to one or the other of the previous types of variable..

\paragraph*{Discrete Variables}
\label{par:discVar}
If $x$ is a discrete variable, taking values in an alphabet $\{x_i\}_{1\leq i \leq Q}$ then the maximum entropy probability distribution $P(x_i) = Pr(x = x_i)$ reads:

\begin{equation}
P(x_i) =\frac{1}{\Zcal(\lambda_1, ...,\lambda_{N_c})} \, \exp{\left( \sum_j \lambda_j f_j(x_i) \right)}
\label{maxentDistDisc}
\end{equation}

where
\begin{equation}
\Zcal(\lambda_1, ...,\lambda_{N_c})= \sum_{i=1}^{Q} \exp{\left( \sum_j \lambda_j f_j(x_i) \right)}
\label{partFuncDisc}.
\end{equation}

\paragraph*{Continuous Variables}
\label{par:contVar}
On the other hand, if $y$ is a real valued variable, one can show that the probability density function $p(y)$ reads 

\begin{equation}
p(y) =\frac{1}{\Zcal(\lambda_1, ...,\lambda_{N_c})} \, \exp{\left( \sum_j \lambda_j f_j(y) \right)}
\label{maxentDistCont}
\end{equation}

where
\begin{equation}
\Zcal(\lambda_1, ...,\lambda_{N_c})= \int_{p(y)>0} \exp{\left( \sum_j \lambda_j f_j(y) \right)} dy.
\label{partFuncDisc}
\end{equation}

In both cases, $\Zcal(\lambda_1, ...,\lambda_{N_c})$ is called partition function and consists in a normalization constant. 

\begin{figure}[!htb]

\noindent\begin{minipage}{\textwidth}

\begin{minipage}{0.47\textwidth}
\centering

\resizebox{0.9\linewidth}{!}{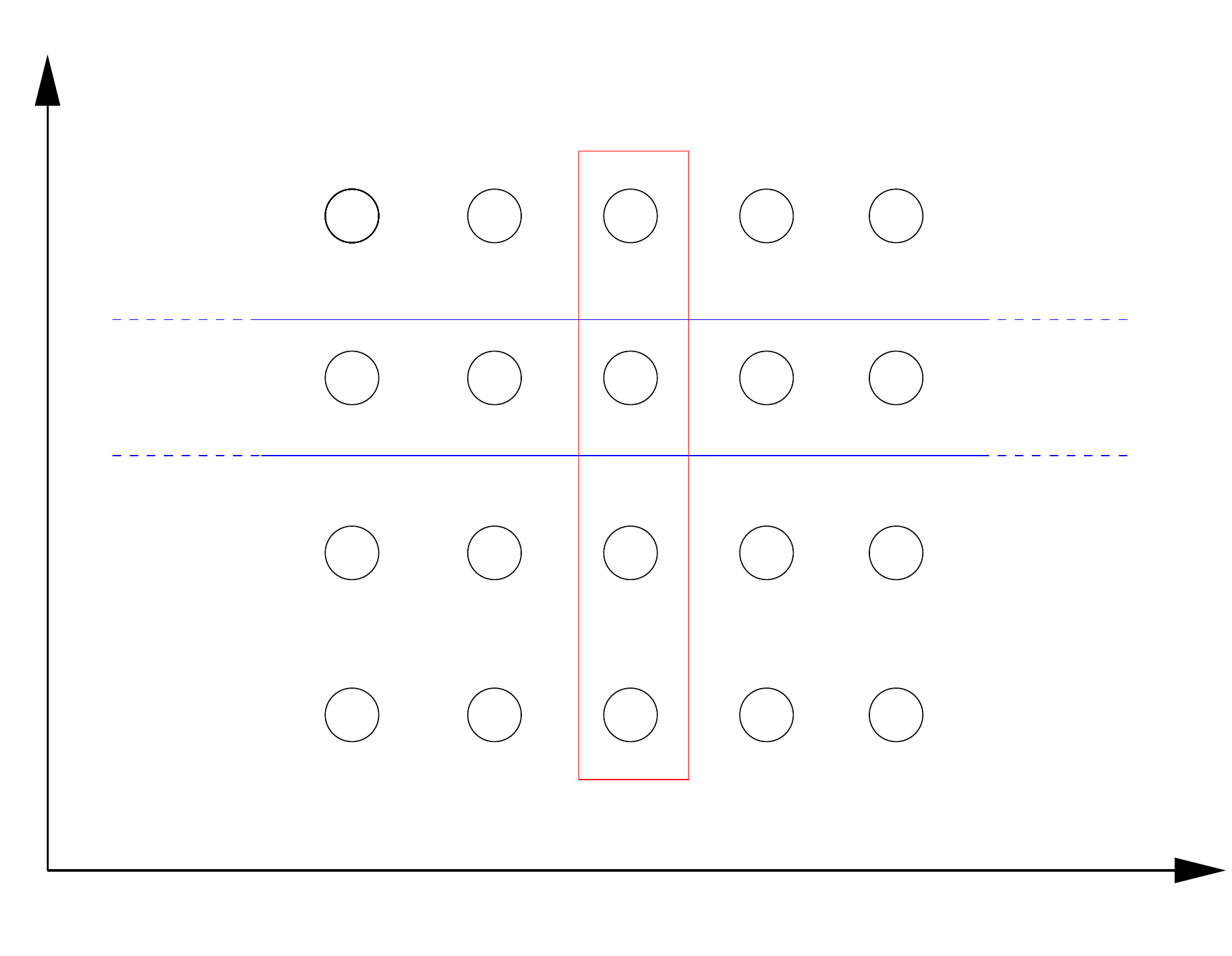}
   \caption{{\bf Schematic view of a sequence.} Graphical representation of a musical sequence of individual notes (red rectangle). Each line represent a voice (blue rectangle), or a specific observable to be involved in the model: microtiming deviations, loudness, metrical position in the bar, etc... }
\label{fig:seq}
\end{minipage}
\hfill
\begin{minipage}{0.06\textwidth}

\end{minipage}
\hfill
\begin{minipage}{0.47\textwidth}
\centering
 \includegraphics[width=0.70\linewidth]{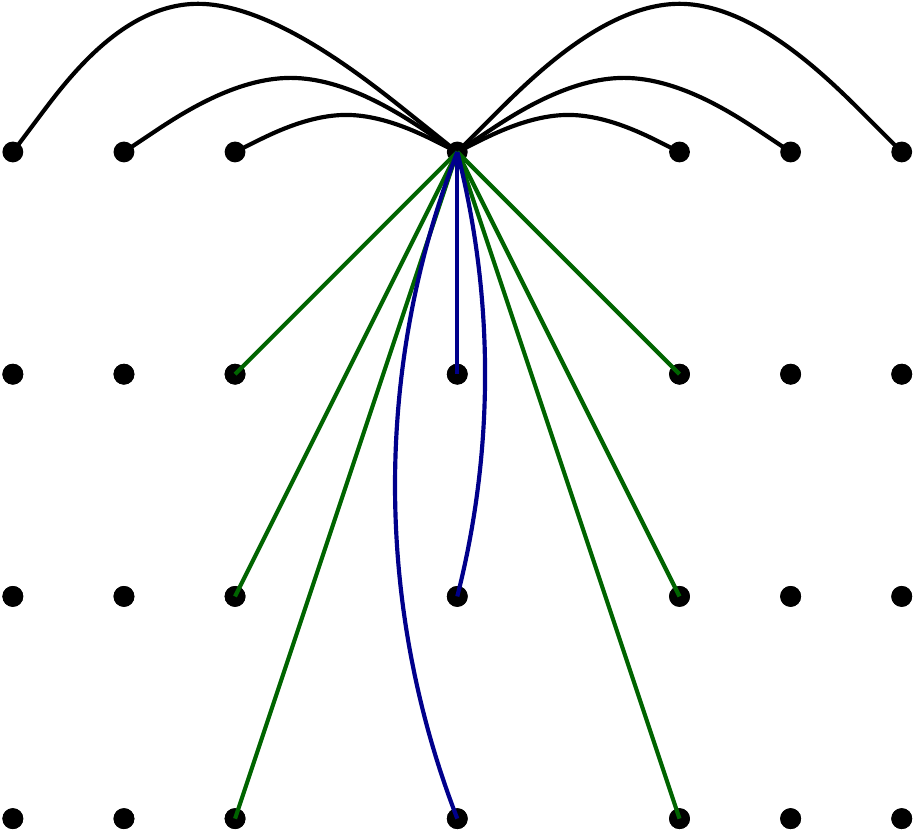}
   \caption{{\bf The Graph Representation.} Graph representing the binary connections ($J_v$) involved in a model associated to the upper voice $v$. Connections can be horizontal (black edges, associated with the same voice's variables), vertical (blue edges, associated to different voice's variables but for the same note), or diagonal (green edges, associated to different voice's variables but for different notes). }
\label{fig:graph}
\end{minipage}
\end{minipage}

\end{figure}

We consider now a sequence of $N$ notes, each of which carrying $N_{cont}$ continuous variables(loudness, microtiming deviations) and $N_{disc}$ discrete $X_{v_{disc}}(n)|_{1\leq n \leq N}$ variables (metrical position in the bar). Transversally, we can see the same sequence as $N_{voices} =  N_{cont} + N_{disc}$ sequences of $N$ elements, all of them being whether discrete or continuous numbers. In this view, we will talk about a superposition of voices (the horizontal blue rectangles on Fig.~\ref{fig:seq}), whereas notes are represented vertically (vertical red rectangles).

Let us denote $Y_{v}$ and $X_v$ respectively a continuous and a discrete variable belonging to the $v$-th voice, and $Y_v(n)$ and $X_v(n)$ respectively their values at the $n$-th note. For each voice $v$, there is a MaxEnt model given by the probability distribution of central variable $Z_v(n)$, conditionned on the values of the neighbouring variables. We define the neighbourhood $\partial Z_v(n)$ of a variable $Z_v$ of the $n$-th note by itself, the variables surrounding $Z_v(n)$ on the same voice ($Z_v(n \pm k)$ with $1\leq k \leq K^{\text{hor}}_\text{max}$), the variables on the same vertical line ($Z_{v'}(n)$  with $v' \neq v$), and the variables diagonally connected to it ($Z_{v'}(n \pm k)$ with $1\leq k \leq K^{\text{diag}}_\text{max}$). For example, the neighbourhood of a variable is represented on Figure~\ref{fig:graph}, with $K^{\text{hor}}_\text{max}=3$ and $K^{\text{diag}}_\text{max}=1$. The expression of the MaxEnt probability (or probability density, for continuous variables) distribution can be then written as follows: 

\begin{eqnarray}
\label{maxEntProb}
P_v(Z_v| \partial Z_v) &=& \frac{1}{\Zcal( \partial Z_v)} \, \exp{\left(- \Hcal_v(Z_v,\partial Z_v) \right)} \\
\label{eq:hamiltonian}
\Hcal(Z_v,\partial Z_v) &=& h_v(Z_v) + \sum_{Z' \in \partial Z_v} J_{Z, Z'}
\end{eqnarray}

where $z$ correspond to the value of variable $Z$. In the previous expressions~\ref{maxEntProb} and~\ref{eq:hamiltonian},, the $J_{Z,Z'}$'s are called the couplings (or interaction energy) between two variables $Z$ and $Z'$, and the $h_v$'s the local fields (or bias). Importantly we consider here only symmetric couplings $ J_{Z,Z'} = J_{Z',Z}$. This terminology comes from statistical physics in which a hamiltonian $\Hcal(Z_v(n),\partial Z_v(n))$ represents an energy associated to a specific configuration of $Z_v(n)$ and $\partial Z_v(n)$. Depending on the nature of the concerned variables $Z$ and $Z'$, $J_{Z,Z'}$ can whether be a real number, a vector or a matrix. 

\begin{eqnarray}
\label{condDisc}
Pr(X_v(n)=x_i| \partial X_v(n)) = P_i(X_v(n)| \partial X_v(n)) &=& \frac{1}{\Zcal( \partial X_v(n))} \, \exp{\left(- h_v(x_i) - \sum_{Z \in \partial X_v(n)} J_{X_v, Z}(x_i,z) \right)} \\
\label{condCont}
Pr(y<Y_v(n)<y+dy | \partial Y_v(n)) = p_v(y| \partial y_v(n)) dy &=&  \frac{1}{\Zcal( \partial Y_v(n))} \, \exp{\left(- h_v \, y - \sum_{Z \in \partial Y_v(n)} J_{X_v, Z}(y, z) \right)} dy 
\end{eqnarray}

Importantly, due to the choice of observables we made, the self-interaction term $J_{Y,Y}$ for continuous variables $Y$ is proportional to $y^2$, taking into account the standard deviation~\ref{eq:ContSD}, while it is zero for integer variables $J_{X,X}=0$. The rest of the couplings to the concerned continuous variable $Y$ contribute to a linear term (proportional to $y$). Consequently, the conditional probability density  $p_v(y| \partial y_v(n))$ is always a gaussian. 

\subsection{Learning the parameters}
\label{sec:learning}

As mentionned in the introduction, we chose to focus on a \emph{translation-invariant} model for simplicity, i.e., for each voice, the elementary module carrying the parameters of the model, represented on figure~\ref{fig:graph}, is translated horizontally for every note of the voice. Every possible position of the graph on the sequence, gives rise to a sample to be used in the inference of the parameters.

\paragraph*{Pseudo log-likelihood}

We use a \emph{Maximum Likelihood Estimation} (MLE) of the model parameters, i.e., we want to optimize the parameters of a model $P_v$ such that the probability to generate the dataset $\Dcal_v$ associated to the voice $v$ using $P_v$ is maximal. The pseudo log-likelihood approximation \cite{ravikumar2010high,ekerberg2}, consists in substituting the probability of the whole sequence $\mathbf{z} = \{ Z_v(n)\}_{1\leq n \leq N}$ by the product of the conditional probabilities of each sample $\mathbf{s_v} = \{ s_v(m)\}_{1\leq m \leq M}$ where $M = N - 2 \text{max}( K^{\text{hor}}_\text{max}, K^{\text{diag}}_\text{max})$ is the number of samples of the sequence, conditionned on the values of the neighbouring variables. The resulting negative pseudo log-likelihood associated to voice $v$, $\Lcal_v$ reads:

\begin{equation} 
\label{logLikelihood}
\Lcal_v(\{J_v\},h_v|\Dcal_v) = -\frac{1}{M} \sum_{m=1}^M \log P_v(z^m_v | \partial z^m_v) 
\end{equation}

Since all the couplings are symmetric, the vertical and the diagonal ones are involved in two different models, corresponding to the two different voices. The determination of the $N_\text{voices}$ models are then entangled and cannot be done sequentially, which lead us to minimize the sum of the negative log-likelihoods over all the voices, consistently with the pseudo log-likelihood approximation. 

\begin{equation} 
\label{pseudologLikelihood}
\Lcal(\{ \{J_v\},h_v\}_{1 \leq v \leq N_\text{voices}}|\Dcal) = \sum_{v} \Lcal_v(\{J_v\},h_v|\Dcal_v) 
\end{equation}
where $\Dcal$ represents the whole dataset $\bigcup\limits_{v} \Dcal_v$. The minimization of $\Lcal$ in \ref{pseudolofLikelihood} does not generally imply the minimization of the individual $\Lcal_v$, however it gives a good approximation of it while the dataset possesses an overall consistency \cite{ekerberg2}.

\paragraph*{The Corpus} The corpus consists in a series of midi melodies recorded twice\footnote{one for the training, the other for the validation tests.} by a professional piano player, with a click at a selected tempo and a very neutral accompaniement (chords played on the downbeats). We chose about 172 tunes from the LeadSheet DataBase \cite{LSDB} divided in 5 sub-corpora so to have a balance of expression styles, namely swing, latin, ballad, pop, and groove. The full list of tunes can be found in appendix \ref{app:corpus}. For each tune, the pianist was asked to play the melody freely, without removing any note nor adding any extra note. Of course, being familiar with almost all the tunes, he respected their musical context, and their style conventions: swing phrasing for swing, etc ...

The dataset is formed as follows: the first voice corresponds to the rhythmic score: it is described by the metrical position of a note in its bar, which is encoded in an integer. The normalized midi onset deviation $\delta o$, the normalized midi duration deviation $\delta d$, and the reduced midi velocity $\delta v$ (loudness) were then measured for each note, and constitute the continuous voices :
\begin{eqnarray}
\delta o &=& \frac{\text{onset}_\text{perf} - \text{onset}_\text{score}}{\text{duration}_\text{score}}\\
\delta d &=& \frac{\text{duration}_\text{perf} - \text{duration}_\text{score}}{\text{duration}_\text{score}}\\
\delta v &=& \frac{\text{velocity}_\text{perf}}{127}
\end{eqnarray}

Each sub-corpus was used to train a style model.

\subsection{Generating sequences}
\label{sec:generation}

Once the parameters have been infered, one can generate sequences by sampling from distribution~(\ref{maxEntProb}) using the Metropolis Algorithm~\cite{metropolis1953equation}. Every variable of the sequence is initialized randomly. Then, the following step is repeated: randomly select a variable, compute its probability conditioned by its neighbours, given by eq.~(\ref{condDisc}) and eq.~(\ref{condCont}), and draw a new value from this probability distribution. A stationnary regime is reached after a number of Monte Carlo steps of the order of a few times the number of variables $n_\text{var}$ (in practice $\approx 10 \, n_\text{var}$). 

\section{Results}
\label{seq:results}
\subsection{Quantitative validation}
\label{subsec:quantValid}

\begin{figure}[!htb]
\noindent\begin{minipage}{\textwidth}

\begin{minipage}{0.47\textwidth}
  \begin{center}
   \centerline{\includegraphics[width=0.92\columnwidth]{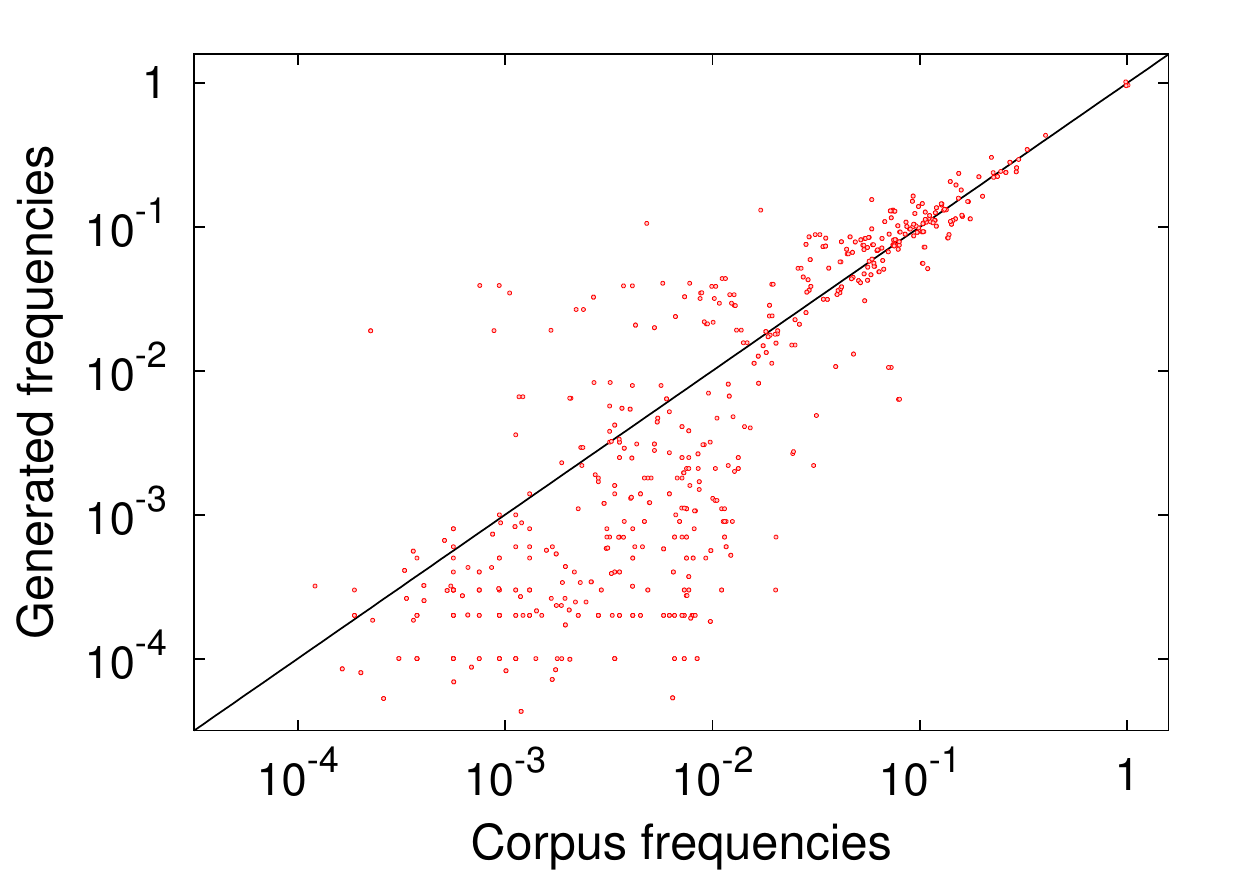}}
  \end{center}
    \caption{{\bf Swing model VS Corpus frequencies:} Scatter plot showing a good agreement between the generated sequence and the corpus. $K^{\text{hor}}_\text{max}=3$ and $K^{\text{diag}}_\text{max}=1$.}
\label{fig:scatter}
\end{minipage}
\hfill
\begin{minipage}{0.02\textwidth}

\end{minipage}
\hfill
\begin{minipage}{0.47\textwidth}
  \begin{center}

   \centerline{\includegraphics[width=1.02\columnwidth]{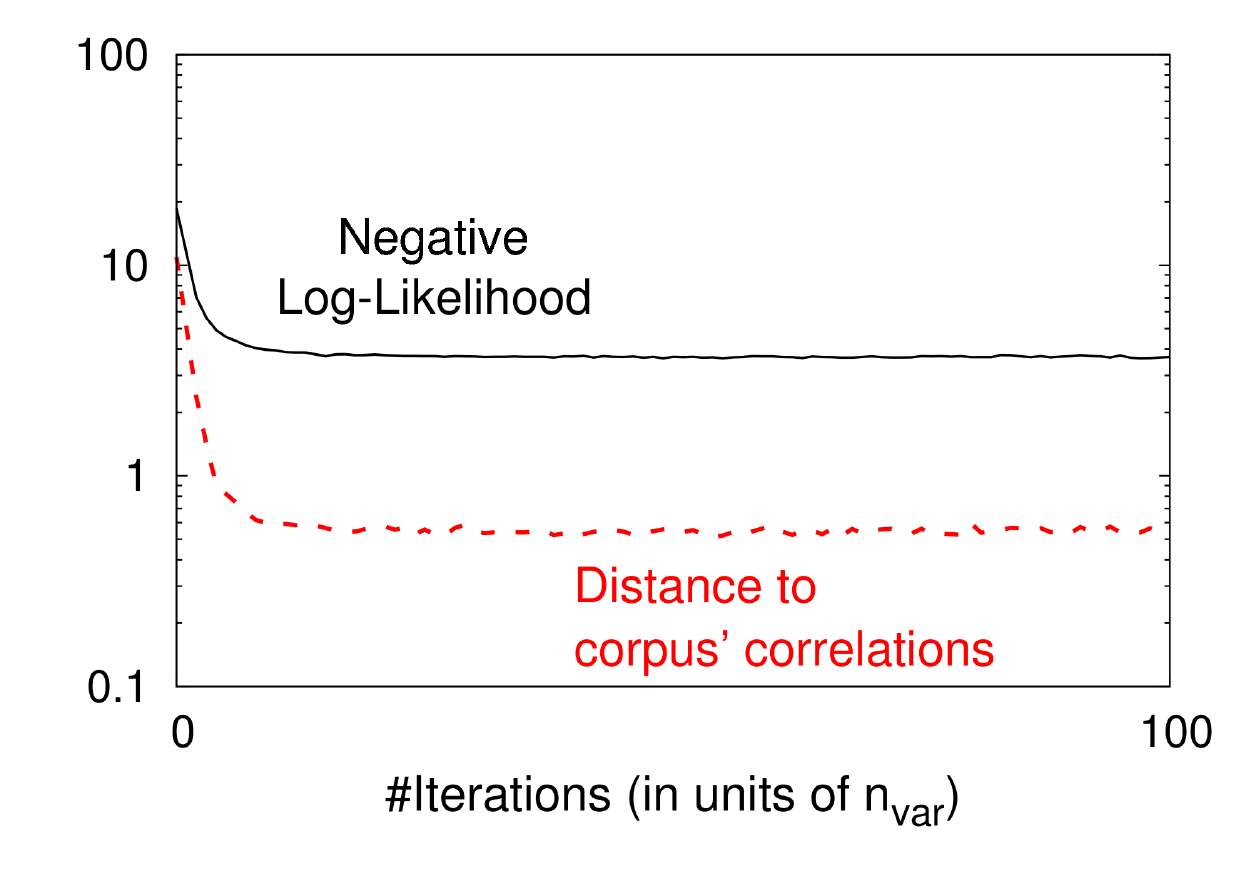}}
  \end{center}
    \caption{{\bf Convergence of the generation:} Convergence appears when both quantities remain almost constant. In practice, about $10 \, n_\text{var}$ iterations are needed.}
\label{fig:generation}
\end{minipage}
\end{minipage}
\end{figure}

In order to test the capacity of our model to capture the structure of correlations in the learning corpus,one can generate a long sequence ($10000$ notes), compute both unary and binary correlations, and compare them to the corpus'. On figure~\ref{fig:scatter}, one can see the a good agreement for high frequencies (more or less $\geq 10^{-2}$), meaning that patterns sufficiently represented in the corpus are well captured by the model. In our example we took $K^{\text{hor}}_\text{max}=3$ and $K^{\text{diag}}_\text{max}=1$, and trained the model on the swing sub-corpus. These values are a good compromise between too many parameters, and a good efficiency in capturing correlations.

Two quantities can be computed along the generation process, in order to monitor the level of convergence of the sequence: The negative log-likelihood, and the distance between the generated sequence correlations, and the corpus' correlations. As we can see on fig.~\ref{fig:generation}, both quantities decrease macroscopically before stabilizing to a stationnary regime, made of typical sequences. We used the same test sequence and the same model (swing) in order to exhibit the convergence properties of the Metropolis procedure.

For each musical style, similar features are observed. A more instructive character of our models is its ability to predict the value of a variable in a sequence that has been played by our performer. This is the role of the test corpus.

\paragraph*{Predictive power:}

We can perform a leave-one-out cross-validation and obtain good results. Let us remind that the coefficient of determination $R^2$ measures the predictive power as follows: $R^2\in[0,1]$, $0$ representing no improved prediction with respect to an empirical gaussian random variable, $1$ representing a perfect prediction. 

\begin{figure}[!htb]
\begin{center}
\begin{tabular}{ |p{3cm}||p{3cm}|p{3cm}|p{3cm}|  }
\hline
Model Name & Onset deviation & Duration deviation  & Loudness\\
 \hline
 \hline
Swing & 0.44 & 0.33 & 0.24 \\
Pop & 0.27 & 0.09 & 0.23\\
Latin & 0.33 & 0.14 & 0.15\\
Groove & 0.34 & 0.11 &  0.21\\
Ballad & 0.32 & 0.26 & 0.20\\
 \hline
\end{tabular}
\label{fig:pred}
\end{center}
\caption{Predictive power of the different models averaged on the test sub-corpora. For each tune of each musical style, we perform a leave-one-out cross-validation procedure, compute the coefficient of determination $R^2$, and give here its average. }
\end{figure}

Generally speaking, the onset deviation seems to be better modeled, whereas the loudness gets worse results than the other observables. Note as well that the swing model has, a better predictive power for any observable, than any other model. We will enter into more details in the discussion section~\ref{sec:conclusion}.

\subsection{Perceptive validation}
\label{subsec:percValid}

In the previous section, we have shown different mathematical criteria which provide the validation of the structure and the implementation of the MaxEnt models for musical expression. To be consistent with our goal, it is necessary to perform a perceptive validation, and this is how we proceeded: 

\paragraph*{Protocol:} We set up a pairwise comparison test of 4 different versions of a melody: our baseline version called ``straight" (with zero microtiming deviations and a constant loudness), a version played by the professional performer (called ``human"), a version generated by our models (called ``MaxEnt") and a control version, called ``random", where microtiming deviations and loudness were sampled with an empirical gaussian distribution\footnote{The same distribution corresponding to zero cross validation coefficient of determination in the previous section}. 
Every participant was asked to listen to two different versions of the melody for each of the 6 tunes we picked. The pair of versions were randomly picked so that after an important number of participations, every version had been compared to every other version the same number of times. The participants were then asked to answer two questions:
\begin{itemize}
\item What version did you prefer ?
\item What version did you find more expressive ?
\end{itemize}

For equity purposes, all four versions were rendered with the same method, and with the same average loudness value. The random and MaxEnt versions were generated with a constrained Metropolis algorithm,\textit{i.e.} performing the Metropolis algorithm in which the voice concerning the rhythmic score is fixed to the score's. Furthermore, the accompaniement was rendered with another soundfont, so it could not be confused with the melody. Finally, a participant could do the test only once.

The following table shows the melodies we choose, with the associated MaxEnt model:

\begin{figure}[!htb]
\begin{center}

\begin{tabular}{|c||c|c|} 
   \hline
Model & Melody & Author \\
    \hline
    \hline
Swing & Donna Lee & C. Parker\\
    \cline{2-2}
 & Blues for Alice & \\ 
    \cline{2-3}
& Bemsha Swing & T. Monk \& D. Best\\
    \hline
Ballad & Central Park West & J. Coltrane\\
\hline
Latin & Mas que nada & J. Ben Jor\\
    \cline{2-3}
 & Meditation & A. C. Jobim\\
\hline
\end{tabular}\label{fig:testList}
\end{center}
\caption{List of the melodies we used for the listening test, with their associated MaxEnt model.}
\end{figure}

A wide communication was done in order to have as many people as possible do the test. We did not aim at selecting any specific category of people. The website hosting the test has been active for 6 weeks during which it collected 244 participations. The musical samples are available at \href{http://ns2292021.ovh.net:3000/}{http://ns2292021.ovh.net:3000/}.

\paragraph*{Intepretation of the results:}

In order to give a general ranking over the four versions, we used a so called Bradley-Terry model \cite{bradley, zermelo} which infers potential $\beta_i$ for each version $i$ consistent with a probabilistic model giving the probability that version $i$ ``beats" version $j$ ($i>j$) :

\begin{equation}
\label{bradTerryModel}
P(i>j)= \frac{e^{\beta_i}}{e^{\beta_i}+e^{\beta_j}}
\end{equation}

\begin{figure}[!htb]
  \begin{center}
   \centerline{\includegraphics[width=0.52\columnwidth]{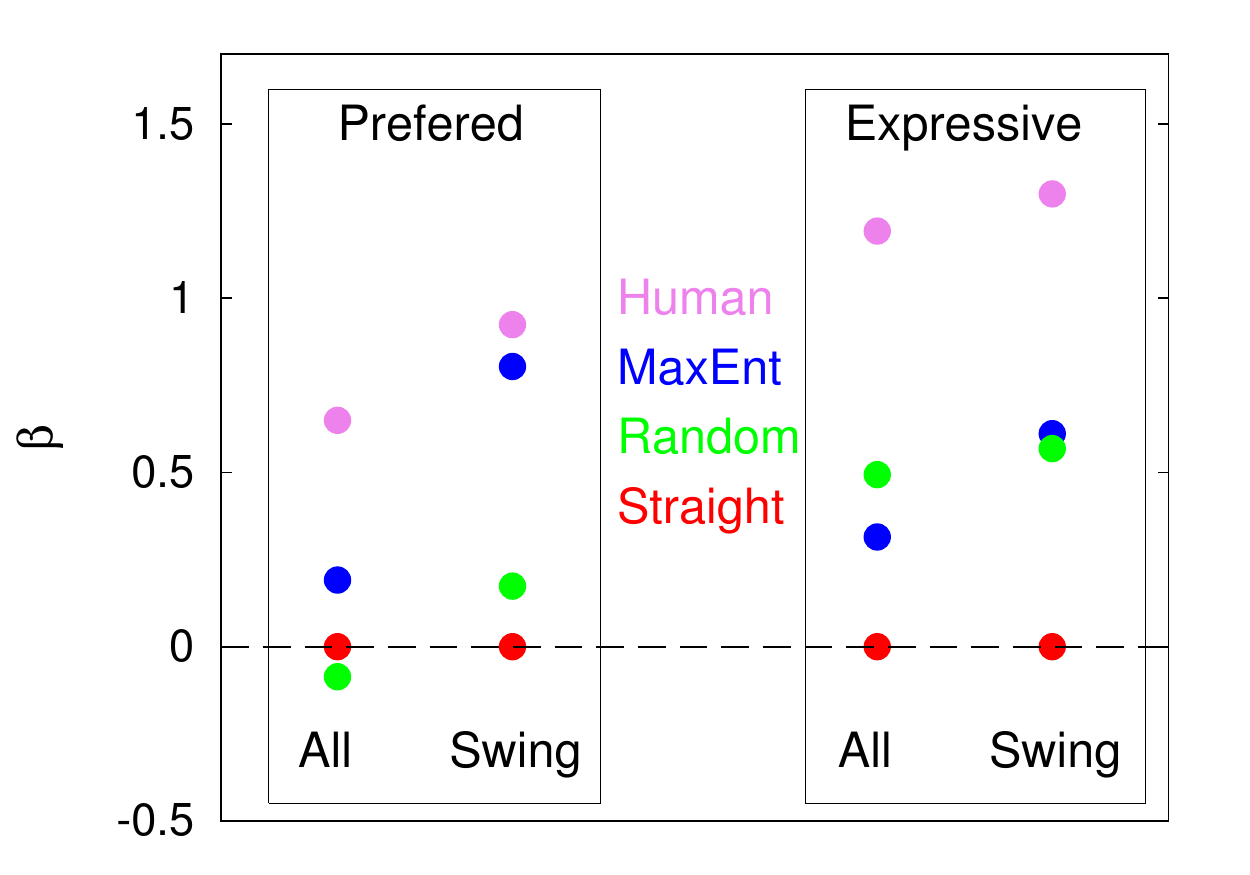}}
  \end{center}
    \caption{{\bf Results of the listening test:}. The left (resp. right) panel shows the version potentials for the first (resp. second) question, \textit{i.e.} about the prefered (resp. more expressive) version. In both cases, we observed an important difference between the results coming from the swing tunes, and the other ones, and this is the reason for which we show two columns, one for the 3 swing tunes, and another for the whole test.}
\label{fig:comp}
\end{figure}

Denoting $n_{i,j}$ the number of votes $i>j$, we use  $P(i>j)=  \frac{n_{i,j}}{n_{i,j}+n_{j,i}}$ as input of the inference algorithm, we impose the potential associated to the straight version to be zero. The results are plotted on figure~\ref{fig:comp}.

Note that any set of values of $P(i>j)$ does not necessarily lead to a fully ordered set of $\beta$. In the four cases shown on figure~\ref{fig:comp}, the algorithm convergences very quickly, ensuring that the votes are very well fitted by the Bradley-Terry model. 

The overall trend puts the human performance in the first position question-wise. The MaxEnt melodies are secondly-prefered, with an significant margin with respect to the other versions. This is particularly the case for the swing, in which there are some strongly defined phrasing codes:  delaying the onset and stressing every even eighth of a bar (upbeats), and increasing the duration of every odd eighth of a bar (downbeats). This pattern is very typical, easy to capture by our MaxEnt model, and strongly recognizable by a human ear. On the other hand, expressive phrasing in ballads or a bossa nova tunes is much less stereotyped, and have an important versatiliy, observed in the data. In particular for slow tempo melodies, a slightly too stressed, or slightly too delayed or anticipated note can sound very artificial. It is very likely that the latter reason explains the important differences between the swing results, with the overall trend.

To the question ``which melody to you find more expressive?", we observe a different tendency: There is a clear domination of the human performance and a clear inferiority of the straight version. In between, both generated versions get comparable scores. It can be explained by the fact that the extreme versions are very recognizable, and neither random nor MaxEnt models success in generating ``as much expressivity" as the one perceived in a human performance. In the title of the question, the notion of ``more expressive" is confusing, and can be understood as "which version differs the most from the straight melody?". In this sense, the results are consistent with the constructions of the versions, but not very instructive from a perceptive point of view.

\section{Conclusion \& perspectives}
\label{sec:conclusion}

We proposed a Maximum Entropy model which captures pairwise correlations between rhythmic score, micro timing and dynamics observables in musical sequences. The model is used to generate new expression sequences that mimic the interpretation of a musician, given musical style. The structure of this model (see Fig.~\ref{fig:graph}) leads to the replication of complex patterns, despite the pairwise nature of the information used. The pairwise correlations have Moreover, our model deals with both continuous and categorical variables, remaining in the context of Maximum Entropy models. 

We performed two different validation processes, quantitative and perceptive, whose conclusions are consistent: MaxEnt models success to capture the local texture of different expression styles. The results are very satisfying in particular for a stereotyped style like swing, whereas they are encouraging for more complex styles, like ballad, or latin jazz. 

One of the most challenging perspective is to allow a model of expression to add or remove notes. This model would be very meaningful in from a musical point of view, but would also cost us a deep modification of the modelization. One should also take into account harmony and pitches, keeping a tractable sample complexity thanks to the MaxEnt model.

\section*{Appendices}

\subsection{Appendix: Model details\\}

In this appendix, we expose the target observables that we want to reproduce. First, the statistical mean and standard deviation of a variable $Y_v$ associated to a continuous voice $v$:
\begin{eqnarray}
\label{eq:ContMean}
f^{\text{cont}}_\text{mean}(Y_v) &=& \frac{1}{N}\sum_{n=1}^N Y_v(n) = \bar{Y_v}\\
\label{eq:ContSD}
f^{\text{cont}}_\text{sd}(Y_v) &=& \sqrt{ \frac{1}{N}\sum_{n=1}^N ( Y_v(n) - \bar{Y_v})^2}.
\end{eqnarray} 
Similarly, for discrete variables, we have the frequency of $x_i$:
\begin{equation}
\label{eq:DiscMean}
f^{\text{disc}}_\text{mean}(X_v)[x_i] = \frac{1}{N}\sum_{n=1}^N \delta_{X_v(n),x_i},
\end{equation}
where $\delta_{.,.}$ is the Kronecker symbol.

Now, we add pairwise correlations between the same variable (on the same voice $v$) separated by $k$ notes are called horizontal correlations, and read :
\begin{eqnarray}
f^\text{cont}_{\text{hor},k}(Y_v) &=& \frac{1}{N-k}\sum_{n=1}^{N-k} Y_v(n) \, Y_v(n+k) \\
f^\text{disc}_{\text{hor},k}(X_v)[x_i, x_j] &=& \frac{1}{N-k}\sum_{n=1}^{N-k} \delta_{X_v(n),x_i} \, \delta_{X_v(n+k),x_j}
\end{eqnarray} 
respectively for a continuous and a discrete variable. Following the same graphical interpretation, vertical correlations exist between variables belonging to voices $v_a$ and $v_b$, and are defined by :
\begin{eqnarray}
f^{\text{cc}}_{\text{vert}}(Y_{v_a},Y_{v_b}) &=& \frac{1}{N}\sum_{n=1}^{N} Y_{v_a}(n)\, Y_{v_b}(n) \\
f^{\text{dd}}_{\text{vert}}(X_{v_a},X_{v_b})[x_i, x_j] &=& \frac{1}{N}\sum_{n=1}^{N} \delta_{X_{v_a}(n),x_i} \, \delta_{X_{v_b}(n),x_j} \\
f^{\text{cd}}_{\text{vert}}(Y_{v_a},X_{v_b})[x_j] &=& \frac{1}{N}\sum_{n=1}^{N} Y_{v_a}(n) \, \delta_{X_{v_b}(n),x_j} \\
f^{\text{dc}}_{\text{vert}}(X_{v_a},Y_{v_b})[x_i] &=& \frac{1}{N}\sum_{n=1}^{N} \delta_{X_{v_a}(n),x_i} \, Y_{v_b}(n)
\end{eqnarray} 
respectively for continuous-continuous, discrete-discrete, or continuous-discrete variable pairs. Finally, the diagonal correlations can be written very similarly : 
\begin{eqnarray}
f^{\text{cc}}_{\text{diag},k}(Y_{v_a},Y_{v_b}) &=& \frac{1}{N-k}\sum_{n=1}^{N-k} Y_{v_a}(n) \, Y_{v_b}(n+k) \\
f^{\text{dd}}_{\text{diag},k}(X_{v_a},X_{v_b})[x_i, x_j] &=& \frac{1}{N-k}\sum_{n=1}^{N-k} \delta_{X_{v_a}(n),x_i}\, \delta_{X_{v_b}(n+k),x_j} \\
f^{\text{cd}}_{\text{diag},k}(Y_{v_a},X_{v_b})[x_j] &=& \frac{1}{N-k}\sum_{n=1}^{N-k} Y_{v_a}(n) \, \delta_{X_{v_b}(n+k),x_j}\\
f^{\text{dc}}_{\text{diag},k}(X_{v_a},Y_{v_b})[x_i] &=& \frac{1}{N-k}\sum_{n=1}^{N-k} \delta_{X_{v_a}(n),x_i}\, Y_{v_b}(n+k).
\end{eqnarray} 
\\

\subsection{Appendix: Corpus\\}
\label{app:corpus}

Yesterday by The Beatles\\
Billie Jean by Michael Jackson\\
Eleanor Rigby by The Beatles\\
Happy by Pharrell Williams\\
All My Loving by The Beatles\\
Eight Days A Week by The Beatles\\
Michelle by The Beatles\\
And I Love Her by The Beatles\\
Folsom Prison Blues by Johnny Cash\\
Fields Of Gold by Sting\\
So What by Miles Davis\\
Old Man by Neil Young\\
Feel by Robbie William\\
Moondance by Van Morrison\\
Sir Duke by Stevie Wonder\\
I Shot The Sheriff by Bob Marley\\
Flamenco Sketches by Miles Davis\\
Hello by Lionel Richie\\
Just The Way You Are by Billy Joel\\
In The Mood by Glenn Miller\\
You Are The Sunshine Of My Life by Stevie Wonder\\
On The Road Again by Willie Nelson\\
Lively Up Yourself by Bob Marley\\
Watermelon Man by Herbie Hancock\\
Both Sides Now by Joni Mitchell\\
Celebration by Kool \& the Gang\\
Naima by John Coltrane\\
Giant Steps by John Coltrane\\
Blue Train by John Coltrane\\
My Life by Billy Joel\\
Cantaloupe Island by Herbie Hancock\\
Chameleon by Herbie Hancock\\
Goodbye Pork Pie Hat by Charles Mingus\\
Cousin Mary by John Coltrane\\
Y.M.C.A by Village People\\
I Feel Good by James Brown\\
Jeru by Miles Davis\\
Maiden Voyage by Herbie Hancock\\
Every Breath You Take by Sting\\
Sunny by Bobby Hebb\\
Strangers Like me by Phil Collins\\
In A Sentimental Mood by Duke Ellington\\
Countdown by John Coltrane\\
Spiral by John Coltrane\\
Self Portrait In Three Colors by Charles Mingus\\
Wave by Antonio Carlos Jobim\\
Locomotion by John Coltrane\\
Caravan by Duke Ellington \& Juan Tizol\\
Boogie Stop Shuffle by Charles Mingus\\
Milestones by Miles Davis\\
Blue Monk by Thelonious Monk\\
Can't Smile Without You by Barry Manilow\\
Nuages by Django Reinhardt \& Jacques Larue\\
Epistrophy by Thelonious Monk\\
Song For My Father by Horace Silver\\
Fables Of Faubus by Charles Mingus\\
Lazybird by John Coltrane\\
Son Of Mr. Green Genes by Frank Zappa\\
Too High by Stevie Wonder\\
Mood Indigo by Duke Ellington\\
Continuum by Jaco Pastorius\\
Jelly Roll by Charles Mingus\\
Pussy Cat Dues by Charles Mingus\\
Part-Time Lover by Stevie Wonder\\
Now's The Time by Charlie Parker\\
All in Love Is Fair by Stevie Wonder\\
Bird Calls by Charles Mingus\\
Acknowledgement (Part 1 of A LOVE SUPREME) by John Coltrane\\
Another Star by Stevie Wonder\\
Strode Rode by Sonny Rollins\\
Chega de Saudade (No More Blues) by Antonio Carlos Jobim\\
Fine And Mellow (My Man Don't Leave Me) by null\\
When You Got A Good Friend by Robert Johnson\\
Triste by Antonio Carlos Jobim\\
Resolution (Part 2 of A LOVE SUPREME) by John Coltrane\\
Evidence by Thelonious Monk\\
Pannonica by Thelonious Monk\\
Central Park West by John Coltrane\\
Sophisticated Lady by Duke Ellington\\
Bemsha Swing by Thelonious Monk\\
Brilliant Corners by Thelonious Monk\\
Solitude by Duke Ellington\\
Dolphin Dance by Herbie Hancock\\
That Girl by Stevie Wonder\\
Four On Six by Wes Montgomery\\
American Patrol by Glenn Miller\\
Yardbird Suite by Charlie Parker\\
In Walked Bud by Thelonious Monk\\
Long Gone Lonesome Blues by Hank Williams\\
Equinox by John Coltrane\\
Stolen Moments by Oliver Nelson\\
Scrapple From The Apple by Charlie Parker\\
Little Brown Jug by Glenn Miller\\
Off Minor by Thelonious Monk\\
Linus And Lucy by Vince Guaraldi\\
Infant Eyes by Wayne Shorter\\
Four by Miles Davis\\
Impressions by John Coltrane\\
Spain by Chick Corea\\
Irene by Caetano Veloso\\
Love In Vain by Robert Johnson\\
Confirmation by Charlie Parker\\
Speak No Evil by Wayne Shorter\\
Mas Que Nada by Jorge Ben\\
Thelonious by Thelonious Monk\\
Witch Hunt by Wayne Shorter\\
Seven Steps To Heaven by Miles Davis\\
A Night In Tunisia by Dizzy Gillespie\\
Chove chuva by Jorge Benjor\\
Qualquer coisa by Caetano Veloso\\
Summertime by George Gershwin \& Ira Gershwin, Du Bose \& Dorothy Heyward\\
Un Poco Loco by Bud Powell\\
Friends To Go by Paul Mc Cartney\\
Sgt Peppers Lonely Hearts Club Band by The Beatles\\
Butterfly by Herbie Hancock\\
Donna Lee by Charlie Parker\\
Queixa by Caetano Veloso\\
Look To The Sky by Antonio Carlos Jobim\\
Straight, No Chaser by Thelonious Monk\\
Ran Kan Kan by Tito Puente\\
Menino do Rio by Caetano Veloso\\
Search For Peace by McCoy Tyner\\
Blues On The Corner by McCoy Tyner\\
Nefertiti by Miles Davis\\
Daphne by Django Reinhardt\\
Crystal Silence by Chick Corea\\
Petite Fleur (Little Flower) by Sidney Bechet\\
Meditation by Antonio Carlos Jobim\\
Blues for Alice by Charlie Parker\\
Antigua by Antonio Carlos Jobim\\
Beleza pura by Caetano Veloso\\
Blue In Green by Bill Evans\\
Tell Me A Bedtime Story by Herbie Hancock\\
Speak Like A Child by Herbie Hancock\\
Luz do Sol by Caetano Veloso\\
Foi um rio que passou em minha vida by Paulinho da viola\\
Diminushing by Django Reinhardt\\
Vivo sonhando by Tom Jobim\\
Aguas de mar\c co by Tom Jobim\\
Choro by Tom Jobim\\
Favela by Antonio Carlos Jobim\\
Falando de amor by Tom Jobim\\
One Note Samba by Antonio Carlos Jobim\\
Retrato Em Branco E Preto by Antonio Carlos Jobim\\
Ligia by Antonio Carlos Jobim\\
Pais tropical by Jorge Benjor\\
Aquele abra\c co by Gilberto Gil\\
Samba do avi\~{a}o by Tom Jobim\\
Trilhos Urbanos by Caetano Veloso\\
Lullaby Of Birdland by George Gershwin\\
Voce E Linda by Caetano Veloso\\
Back in Bahia by Gilberto Gil\\
Samba Cantina by Paul Desmond\\
I Got You (I Feel Good) by James Brown\\
Minha Saudade by Joao Donato\\
Solar by Miles Davis\\

\section*{Acknowledgments}
This research is conducted within the Lrn2Cre8 project which received funding from the European Union's Seventh Framework Programme (FET grant agreement n. 610859). The authors thank Jason Sakellariou, Ga\"etan Hadjeres, and Maarten Grachten for fruitful discussions.

\bibliography{biblio}

\end{document}

%% file: sequence.pdf_t
\begin{picture}(0,0)%
\includegraphics{sequence.pdf}%
\end{picture}%
\setlength{\unitlength}{4144sp}%
\begingroup\makeatletter\ifx\SetFigFont\undefined%
\gdef\SetFigFont#1#2#3#4#5{%
  \reset@font\fontsize{#1}{#2pt}%
  \fontfamily{#3}\fontseries{#4}\fontshape{#5}%
  \selectfont}%
\fi\endgroup%
\begin{picture}(8542,6627)(346,-6052)
\put(361,344){\makebox(0,0)[lb]{\smash{{\SetFigFont{17}{20.4}{\familydefault}{\mddefault}{\updefault}{\color[rgb]{0,0,0}Voices ($n_v$)}%
}}}}
\put(8371,-5956){\makebox(0,0)[lb]{\smash{{\SetFigFont{17}{20.4}{\familydefault}{\mddefault}{\updefault}{\color[rgb]{0,0,0}Time ($n$)}%
}}}}
\put(3466,-5146){\makebox(0,0)[lb]{\smash{{\SetFigFont{17}{20.4}{\familydefault}{\mddefault}{\updefault}{\color[rgb]{0,0,0}$n-1$}%
}}}}
\put(5356,-5146){\makebox(0,0)[lb]{\smash{{\SetFigFont{17}{20.4}{\familydefault}{\mddefault}{\updefault}{\color[rgb]{0,0,0}$n+1$}%
}}}}
\put(6301,-5146){\makebox(0,0)[lb]{\smash{{\SetFigFont{17}{20.4}{\familydefault}{\mddefault}{\updefault}{\color[rgb]{0,0,0}$n+2$}%
}}}}
\put(2521,-5146){\makebox(0,0)[lb]{\smash{{\SetFigFont{17}{20.4}{\familydefault}{\mddefault}{\updefault}{\color[rgb]{0,0,0}$n-2$}%
}}}}
\put(4636,-5146){\makebox(0,0)[lb]{\smash{{\SetFigFont{17}{20.4}{\familydefault}{\mddefault}{\updefault}{\color[rgb]{0,0,0}$n$}%
}}}}
\put(1261,-1006){\makebox(0,0)[lb]{\smash{{\SetFigFont{17}{20.4}{\familydefault}{\mddefault}{\updefault}{\color[rgb]{0,0,0}$v_4$}%
}}}}
\put(1261,-2176){\makebox(0,0)[lb]{\smash{{\SetFigFont{17}{20.4}{\familydefault}{\mddefault}{\updefault}{\color[rgb]{0,0,0}$v_3$}%
}}}}
\put(1261,-4471){\makebox(0,0)[lb]{\smash{{\SetFigFont{17}{20.4}{\familydefault}{\mddefault}{\updefault}{\color[rgb]{0,0,0}$v_1$}%
}}}}
\put(1216,-3391){\makebox(0,0)[lb]{\smash{{\SetFigFont{17}{20.4}{\familydefault}{\mddefault}{\updefault}{\color[rgb]{0,0,0}$v_2$}%
}}}}
\end{picture}%